___

# Hand geometry based recognition with a MLP classifier

Marcos Faundez-Zanuy[1], Miguel A. Ferrer-Ballester[2], Carlos M. Travieso-González[2], Virginia Espinosa-Duro[1]

[1] Escola Universitària Politècnica de Mataró (UPC), Barcelona, Spain
{faundez, espinosa}@eupmt.es
http://www.eupmt.es/veu

[2] Dpto. de Señales y Comunicaciones, Universidad de Las Palmas de Gran Canaria
Campus de Tafira, E-35017, Las Palmas de Gran Canaria, Spain
{mferrer, ctravieso}@dsc.ulpgc.es
http://www.gpds.ulpgc.es

**Abstract.** This paper presents a biometric recognition system based on hand geometry. We describe a database specially collected for research purposes, which consists of 50 people and 10 different acquisitions of the right hand. This database can be freely downloaded. In addition, we describe a feature extraction procedure and we obtain experimental results using different classification strategies based on Multi Layer Perceptrons (MLP). We have evaluated identification rates and Detection Cost Function (DCF) values for verification applications. Experimental results reveal up to 100% identification and 0% DCF.

## 1 Introduction

In recent years, hand geometry has become a very popular biometric access control, which has captured almost a quarter of the physical access control market [1]. Even if the fingerprint [2], [3] is the most popular access system, the study of other biometric systems is interesting, because the vulnerability of a biometric system [4] can be improved using some kind of data fusion [5] between different biometric traits. This is a key point in order to popularize biometric systems [6], in addition to privacy issues [7].

Although some commercial systems rely on a three-dimensional profile of the hand, in this paper we study a system based on two dimensional profiles. Even though three dimensional devices provide more information than two dimensional ones, they require a more expensive and voluminous hardware.

A two-dimensional profile of a hand can be get using a simple document scanner, which can be purchased for less than 100 USD. Another possibility is the use of a digital camera, whose cost is being dramatically reduced in the last years.

In our system, we have decided to use a conventional scanner instead of a digital photo camera, because it is easier to operate, and cheaper. This paper can be summarized in three main parts: section two describes a database which has been specially acquired for this work. In section three, we describe the pre-processing and feature extraction. Section four provides experimental results on identification and verification rates using neural net classifiers. Finally, conclusions are summarized.

## 2   Database Description

Our database consists of 10 different acquisitions of 50 people, acquiring the right hand of each user. We have used a conventional document scanner, where the user can place the hand palm freely over the scanning surface; we do not use pegs, templates or any other annoying method for the users to capture their hands [8]. The images have been acquired with a typical desk-scanner using 8 bits per pixel (256 gray levels), and a resolution of 150dpi. To facilitate later computation, every scanned image has been scaled by a factor of 20%.

## 3   Feature Extraction

This stage can be split into the following steps: binarization, contour extraction, work out the geometric measurements, and finally the features are stored in a new reduced database (see figure 1)

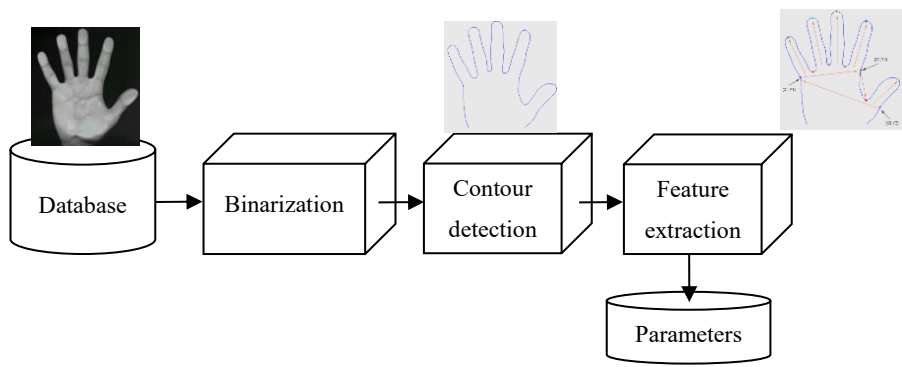

**Fig. 1.** Steps in the feature extraction process

*Step 1*. The goal of this step is the conversion from a 8 bit per pixel image to a monochrome image (1 bit per pixel). As the contrast between the image and the background

______________________________________________________________

is quite high, it reduces the complexity of the binarization process. After several experiments changing the threshold and evaluating the results with different images extracted from the database, we reach the conclusion that with a selected threshold of 0.25 the results were adequate for our purposes. We discard to use other binarization algorithm such as the suggested by Lloyd, Ridler-Calvar and Otsu [9] because the results are similar and the computational burden is higher.

*Step 2*. The goal is to find the limits between the hand and the background and obtain a numerical sequence describing the hand-palm shape. Contour following is a procedure by which we run through the hand silhouette by following the image's edge. We have implemented an algorithm, which is a modification of the method created by Sonka, Hlavac and Boyle [10].

*Step3*. Several intermediate steps have been performed to detect the main points of the hands from the image database. The method for the geometric hand-palm features extraction is quite straightforward. From the hand image, we locate the following main points: finger tips, valleys between the fingers and three more points that are necessary to define the hand geometry precisely. Finally, using all the main points previously computed, the geometric measurements are obtained. We take the eight following distances: Length of the 5 fingers, distances between points (X1, Y1) and (X2,Y2), points (X2,Y2) and the valley between the thumb and first finger and the points (X3,Y3) and (X1,Y1). Figure 1 shows the final results along with the geometric measurements taken into account.

## 4   Experimental Results

Biometric systems can be operated in two ways:

- Identification: In this approach no identity is claimed from the person. The automatic system must determine who is trying to access.
- Verification: In this approach the goal of the system is to determine whether the person is who he/she claims to be. This implies that the user must provide an identity and the system just accepts or rejects the users according to a successful or unsuccessful verification. Sometimes this operation mode is named authentication or detection.

For identification, if we have a population of N different people, and a labeled test set, we just need to count the number of identities correctly assigned.

Verification systems can be evaluated using the False Acceptance Rate (FAR, those situations where an impostor is accepted) and the False Rejection Rate (FRR, those situations where a user is incorrectly rejected), also known in detection theory as False Alarm and Miss, respectively. There is trade-off between both errors, which has to be usually established by adjusting a decision threshold. The performance can be plotted

___________________________________________________________________

in a ROC (Receiver Operator Characteristic) or in a DET (Detection error trade-off) plot [11]. DET plot uses a logarithmic scale that expands the extreme parts of the curve, which are the parts that give the most information about the system performance.

In order to summarize the performance of a given system with a single number, we have used the minimum value of the Detection Cost Function (DCF). This parameter is defined as [11]:

$$DCF = C_{miss} \times P_{miss} \times P_{true} + C_{fa} \times P_{fa} \times P_{false} \qquad (1)$$

where $C_{miss}$ is the cost of a miss (rejection), $C_{fa}$ is the cost of a false alarm (acceptance), $P_{true}$ is the a priori probability of the target, and $P_{false} = 1 - P_{true}$. We have used $C_{miss} = C_{fa} = 1$.

***Multi-Layer Perceptron classifier trained in a discriminative mode.*** We have trained a Multi-Layer Perceptron (MLP) [12] as discriminative classifier in the following fashion: when the input data belongs to a genuine person, the output (target of the NNET) is fixed to 1. When the input is an impostor person, the output is fixed to –1. We have used a MLP with 40 neurons in the hidden layer, trained with gradient descent algorithm with momentum and weight/bias learning function. We have trained the neural network for 2500 and 10000 epochs using regularization. We also apply a multi-start algorithm and we provide the mean, standard deviation, and best obtained result for 50 random different initializations. The input signal has been fitted to a [–1, 1] range in each component.

***Error correction codes.*** Error-control coding techniques [13] detect and possibly correct errors that occur when messages are transmitted in a digital communication system. To accomplish this, the encoder transmits not only the information symbols, but also one or more redundant symbols. The decoder uses the redundant symbols to detect and possibly correct whatever errors occurred during transmission.

Block coding is a special case of error-control coding. Block coding techniques map a fixed number of message symbols to a fixed number of code symbols. A block coder treats each block of data independently and is a memoryless device. The information to be encoded consists of a sequence of message symbols and the code that is produced consists of a sequence of codewords. Each block of k message symbols is encoded into a codeword that consists of n symbols; in this context, k is called the message length, n is called the codeword length, and the code is called an [n, k] code.

A message for an [n, k] BCH (Bose-Chaudhuri-Hocquenghem) code must be a k-column binary Galois array. The code that corresponds to that message is an n-column binary Galois array. Each row of these Galois arrays represents one word.

BCH codes use special values of n and k:

- n, the codeword length, is an integer of the form $2m–1$ for some integer m > 2.
- k, the message length, is a positive integer less than n.

___________________________________________________________________

However, only some positive integers less than n are valid choices for k. This code can correct all combinations of t or fewer errors, and the minimum distance between codes is:

$$d_{min} \geq 2t+1 \qquad (2)$$

Table 2 shows some examples of suitable values for BCH codes,

**Table 1.** Examples of values for BCH codes.

| n | 7 | 5 | | | 31 | | | | |
|---|---|---|---|---|----|----|----|----|---|
| k | 4 | 11 | 7 | 5 | 26 | 21 | 16 | 11 | 6 |
| t | 1 | 1 | 2 | 3 | 1 | 2 | 3 | 5 | 7 |

***Multi-class learning problems via error-correction output codes.*** Multi-class learning problems involve finding a definition for an unknown function $f(\vec{x})$ whose range is a discrete set containing $k > 2$ values (i.e. $k$ classes), and $\vec{x}$ is the set of measurements that we want to classify. We must solve the problem of learning a *k*-ary classification function $f : \Re^n \rightarrow \{1, \cdots, k\}$ from examples of the form $\{\vec{x}_i, f(\vec{x}_i)\}$. The standard neural network approach to this problem is to construct a 3-layer feed-forward network with *k* output units, where each output unit designates one of the *k* classes. During training, the output units are clamped to 0.0, except for the unit corresponding to the desired class, which is clamped at 1.0. During classification, a new $\vec{x}$ value is assigned to the class whose output unit has the highest activation. This approach is called [14], [15], [16] the *one-per-class* approach, since one binary output function is learnt for each class.

***Experimental Results.*** We use a Multi-layer perceptron with 10 inputs, and h hidden neurons, both of them with *tansig* nonlinear transfer function. This function is symmetrical around the origin. Thus, we modify the output codes replacing each "0" by "–1". In addition, we normalize the input vectors $\vec{x}$ for zero mean and maximum modulus equal to 1. The computation of Mean Square Error (MSE), and Mean Absolute Difference (MAD) between the obtained output and each of the codewords provides a distance measure.

We have converted this measure into a similarity measure computing (1 – distance). We will summarize the Multi-Layer Perceptron number of neurons in each layer using the following nomenclature: *inputs× hidden× output*. In our experiments, the number of inputs is fixed to 10, and the other parameters can vary according to the selected strategy.

We have evaluated the following strategies (each one has been tested with 5 and 3 hands for training, and the remaining ones for testing):

− One-per-class: 1 MLP 10×40×50 (table 2)
− Natural binary code: 1 MLP 10×40×6 (table 3)
− Error Correction Output Code (ECOC) using BCH (15, 7) (table 4) and BCH (31, 6) (table 5).

______________________________________________________________________

– Error Correction Output Code (ECOC) using random generation (table 6).

**Table 2.** 1 MLP 10×40×50 (one-per-class).

| Epoch | Train=5 hands, test=5 hands | | | | | | Train=3 hands, test=7 hands | | | | | |
|---|---|---|---|---|---|---|---|---|---|---|---|---|
| | Identif. rate (%) | | | Min(DCF) (%) | | | Identif. rate (%) | | | Min(DCF) (%) | | |
| | mean | σ | max | mean | σ | min | mean | σ | max | mean | σ | min |
| 2500 | 98.30 | 0.39 | 98.80 | 0.69 | 0.2 | 0.34 | 97.71 | 0.70 | 99.14 | 0.86 | 0.2 | 0.50 |
| 10000 | 98.23 | 0.47 | 99.2 | 0.67 | 0.16 | 0.37 | 97.79 | 0.64 | 98.57 | 0.86 | 0.18 | 0.53 |

**Table 3.** 1 MLP 10×40×6 (Natural binary code).

| | Epoch | Train=5 hands, test=5 hands | | | | | | Train=3 hands, test=7 hands | | | | | |
|---|---|---|---|---|---|---|---|---|---|---|---|---|---|
| | | Identif. rate (%) | | | Min(DCF) (%) | | | Identif. rate (%) | | | Min(DCF) (%) | | |
| | | mean | σ | max | mean | σ | min | mean | σ | max | mean | σ | min |
| MAD | 2500 | 95.97 | 1.25 | 98.4 | 3.94 | 0.52 | 2.74 | 92.43 | 1.49 | 96.57 | 5.70 | 0.45 | 4.79 |
| MAD | 10000 | 96.42 | 1 | 98.4 | 3.80 | 0.49 | 2.77 | 92.66 | 1.17 | 95.14 | 5.58 | 0.39 | 4.81 |
| MSE | 2500 | 95.97 | 1.25 | 98.4 | 0.88 | 0.3 | 0.38 | 92.43 | 1.49 | 96.57 | 2.60 | 0.41 | 1.87 |
| MSE | 1000 | 96.42 | 1 | 98.4 | 0.83 | 0.3 | 0.29 | 92.66 | 1.17 | 95.14 | 2.53 | 0.42 | 1.60 |

**Table 4.** 1 MLP 10×40×50 (ECOC BCH (31, 6)).

| | Epoch | Train=5 hands, test=5 hands | | | | | | Train=3 hands, test=7 hands | | | | | |
|---|---|---|---|---|---|---|---|---|---|---|---|---|---|
| | | Identif. rate (%) | | | Min(DCF) (%) | | | Identif. rate (%) | | | Min(DCF) (%) | | |
| | | mean | σ | max | mean | σ | min | mean | σ | max | mean | σ | min |
| MAD | 2500 | 99.58 | 0.15 | 100 | 0.04 | 0.05 | 0 | 98.54 | 0.60 | 99.71 | 0.49 | 0.21 | 0.12 |
| MAD | 10000 | 99.62 | 0.21 | 100 | 0.03 | 0.04 | 0 | 98.50 | 0.60 | 99.43 | 0.45 | 0.18 | 0.11 |
| MSE | 2500 | 99.58 | 0.15 | 100 | 0.03 | 0.04 | 0 | 98.59 | 0.57 | 99.71 | 0.46 | 0.20 | 0.06 |
| MSE | 10000 | 99.62 | 0.22 | 100 | 0.02 | 0.03 | 0 | 98.53 | 0.59 | 99.43 | 0.43 | 0.17 | 0.11 |

**Table 5.** 1 MLP 10×40×14 (ECOC BCH (15, 7)).

| | Epoch | Train=5 hands, test=5 hands | | | | | | Train=3 hands, test=7 hands | | | | | |
|---|---|---|---|---|---|---|---|---|---|---|---|---|---|
| | | Identif. rate (%) | | | Min(DCF) (%) | | | Identif. rate (%) | | | Min(DCF) (%) | | |
| | | mean | σ | max | mean | σ | min | mean | σ | max | mean | σ | min |
| MAD | 2500 | 99.58 | 0.15 | 100 | 0.04 | 0.05 | 0 | 98.06 | 0.58 | 99.43 | 0.94 | 0.26 | 0.49 |
| MAD | 10000 | 99.62 | 0.21 | 100 | 0.03 | 0.04 | 0 | 98.30 | 0.58 | 99.43 | 0.85 | 0.26 | 0.44 |
| MSE | 2500 | 99.58 | 0.15 | 100 | 0.03 | 0.04 | 0 | 98.07 | 0.58 | 99.14 | 0.47 | 0.18 | 0.18 |
| MSE | 10000 | 99.61 | 0.22 | 100 | 0.02 | 0.03 | 0 | 98.35 | 0.61 | 99.43 | 0.39 | 0.19 | 0.06 |

**Table 6.** 1 MLP 10×40×50 (random ECOC generation).

| | Epoch | Train=5 hands, test=5 hands | | | | | | Train=3 hands, test=7 hands | | | | | |
|---|---|---|---|---|---|---|---|---|---|---|---|---|---|
| | | Identif. rate (%) | | | Min(DCF) (%) | | | Identif. rate (%) | | | Min(DCF) (%) | | |
| | | mean | σ | max | mean | σ | min | mean | σ | max | mean | σ | min |
| M | 2500 | 99.50 | 0.23 | 100 | 0.26 | 0.12 | 0.01 | 98.09 | 0.78 | 99.71 | 1.22 | 0.38 | 0.41 |

|     | 10000 | 99.58 | 0.23 | 100 | 0.23 | 0.12 | 0     | 98.30 | 0.70 | 99.71 | 1.10 | 0.31 | 0.60 |
| --- | ----- | ----- | ---- | --- | ---- | ---- | ----- | ----- | ---- | ----- | ---- | ---- | ---- |
| MSE | 2500  | 99.50 | 0.21 | 100 | 0.13 | 0.01 | 0.004 | 98.14 | 0.78 | 99.71 | 0.85 | 0.33 | 0.22 |
|     | 10000 | 99.58 | 0.23 | 100 | 0.09 | 0.09 | 0     | 98.32 | 0.67 | 99.71 | 0.71 | 0.32 | 0.09 |

# 5  Conclusions

Taking into account the experimental results, we observe the following conclusions:

– Comparing tables 3 and 4, we observe better performance using the one-per-class approach. We think that is due to the larger number of weights when using the first strategy, which lets to obtain a better classifier. Additionally, we can interpret that the larger hamming distance of one-per-class approach lets to improve the results.
– ECOC lets more flexibility with the MLP architecture, because it has a wide range of possibilities for the number of outputs, given a set of users. In addition, experimental results outperform the one-per-class approach. Comparing tables 5 and 6 we see similar performance. Thus, we prefer BCH (15, 7) because it is simpler.
– Although it is supposed that random generation for ECOC should outperform BCH codes, our experimental results reveal better performance when using the latest ones.
– Our results offers better efficacy than other works with similar database size [17-18].

## Acknowledgement

This work has been partially funded by FEDER and MCYT TIC2003-08382-C05-02

Faundez-Zanuy, M., Ferrer-Ballester, M.A., Travieso-González, C.M., Espinosa-Duro, V. (2005). Hand Geometry Based Recognition with a MLP Classifier. In: Zhang, D., Jain, A.K. (eds) Advances in Biometrics. ICB 2006. Lecture Notes in Computer Science, vol 3832. Springer, Berlin, Heidelberg. https://doi.org/10.1007/11608288_96